\RequirePackage{fix-cm}
\documentclass[twocolumn]{svjour3}          
\smartqed  
\usepackage{graphicx}
\usepackage{times}
\usepackage{epsfig}
\usepackage{epstopdf}
\usepackage{amsmath}
\usepackage{amssymb}
\usepackage[labelfont=bf]{caption}
\usepackage{tabularx}
\usepackage[table, xcdraw, dvipsnames]{xcolor}
\usepackage{algorithm}
\usepackage[noend]{algpseudocode}

\usepackage[section]{placeins}
\usepackage{booktabs}
\usepackage{makecell}
\usepackage{multirow}
\usepackage{mmstyles}
\usepackage{siunitx}
\usepackage{marvosym}
\usepackage[numbers]{natbib}
\definecolor{citecolor}{HTML}{0071bc}
\definecolor{tabhighlight}{HTML}{e5e5e5}
\usepackage[colorlinks,citecolor=citecolor]{hyperref}
\usepackage{bbm}
\usepackage{pifont}
\usepackage{subcaption}
\usepackage{float}

\usepackage[normalem]{ulem}
\definecolor{ForestGreen}{rgb}{0.13, 0.55, 0.13}
\definecolor{Green}{rgb}{0.0, 0.5, 0.0}
\definecolor{green(munsell)}{rgb}{0.0, 0.66, 0.47}
\definecolor{green(ryb)}{rgb}{0.4, 0.69, 0.2}
\definecolor{green(pigment)}{rgb}{0.0, 0.65, 0.31}
\definecolor{GrayXMark}{gray}{0.6}

\graphicspath{{figures/}}
\DeclareGraphicsExtensions{.pdf}

\definecolor{blue-violet}{rgb}{0.54, 0.17, 0.89}
\newcommand{\yk}[1]{\textcolor{black}{#1}}

\newcommand{\cprompt}[1]{{\scriptsize\textit{#1}}}
\newcommand{\cinstruct}[1]{{\scriptsize\textit{#1}}}

\usepackage{wrapfig}
\usepackage{setspace}
\usepackage[utf8]{inputenc} 
\usepackage[T1]{fontenc}    
\usepackage{url}            
\usepackage{booktabs}       
\usepackage{amsfonts}       
\usepackage{nicefrac}       
\usepackage{microtype}      
\usepackage{graphicx} 
\usepackage{tabularx}
\usepackage{amsmath}
\usepackage{xspace}
\usepackage{graphbox}
\usepackage{float}
\usepackage{multirow}
\usepackage{booktabs}
\usepackage{adjustbox}
\usepackage{caption,setspace}

\makeatletter
\def\BState{\State\hskip-\ALG@thistlm}
\makeatother
\definecolor{myGreen}{HTML}{33FF00}
\definecolor{myRed}{HTML}{FF3030}
\definecolor{myGrey}{HTML}{AA5555}
\definecolor{myWhite}{HTML}{FFFFFF}
\definecolor{maroon}{cmyk}{0,0.87,0.68,0.32}
\definecolor{petr}{HTML}{5555FF}
\definecolor{josef}{HTML}{FF3030}

\journalname{Noname}
\begin{document}
\begin{sloppypar}

\title{CrowdMoGen: Event-Driven Collective Human Motion Generation}

\author{
 Yukang Cao$^{*}$
    \
 Xinying Guo$^{*}$
   \
 Mingyuan Zhang
   \
 Haozhe Xie
  \
 Chenyang Gu
  \
 Ziwei Liu\textsuperscript{\Letter}  \vspace{0.2cm}\\
  S-Lab, Nanyang Technological University \\
  $^{*}$ Equal contributions \textsuperscript{\Letter} Corresponding author \\
  \url{https://yukangcao.github.io/CrowdMoGen}
}

\authorrunning{Yukang Cao, Xinying Guo \textit{et al.}} 

\institute{Yukang Cao \at
S-Lab, Nanyang Technological University\\
\email{yukang.cao@ntu.edu.sg}
\and
Xinying Guo \at
S-Lab, Nanyang Technological University\\
\email{gxynoz@gmail.com}
\and
Mingyuan Zhang \at
S-Lab, Nanyang Technological University\\
\email{mingyuan001@e.ntu.edu.sg}
\and
Haozhe Xie \at
S-Lab, Nanyang Technological University\\
\email{haozhe.xie@ntu.edu.sg}
\and
Chenyang Gu \at
S-Lab, Nanyang Technological University\\
\email{rhea1018@outlook.com}
\and
Ziwei Liu \at
S-Lab, Nanyang Technological University\\
\email{ziwei.liu@ntu.edu.sg}
}
\date{Received: date / Accepted: date}

\newcommand{\OM}{CrowdMoGen}
\newcommand{\OMO}{CrowdMoGen }

\maketitle



\begin{abstract}

While recent advances in text-to-motion generation have shown promising results, 
\yk{they typically assume all individuals are grouped as a single unit.}
Scaling these methods to handle larger crowds and ensuring that individuals respond appropriately to specific events remains a significant challenge. 
This is primarily due to the complexities of \yk{scene planning--which involves organizing groups, planning their activities, and coordinating interactions--and controllable motion generation.}
In this paper, we present \textbf{\OMO}, the first zero-shot framework for collective motion generation, which effectively groups individuals and generates event-aligned motion sequences from text prompts.
\textbf{1)} \yk{Being limited by the available datasets for training an effective scene planning module in a supervised manner, we instead propose a \textbf{crowd scene planner} that leverages pre-trained large language models (LLMs) to organize individuals into distinct groups. While LLMs offer high-level guidance for group divisions, they lack the low-level understanding of human motion. To address this, we further propose integrating an SMPL-based joint prior to generate context-appropriate activities, which consists of both joint trajectories and textual descriptions. }
\textbf{2)} Secondly, to incorporate the assigned activities into the generative network, we introduce a \textbf{collective motion generator} that \yk{integrates the activities into a transformer-based network in a joint-wise manner}, maintaining the spatial constraints during the multi-step denoising process.  
Extensive experiments demonstrate that \OMO significantly outperforms previous approaches, delivering realistic, event-driven motion sequences that are spatially coherent. 
As the first framework of \yk{collective motion generation}, \OMO has the potential to advance applications in urban simulation, crowd planning, and other large-scale interactive environments.

\end{abstract}
\section{Introduction}

\begin{figure*}[t]
\vspace{-2em}
  \includegraphics[width=\textwidth]{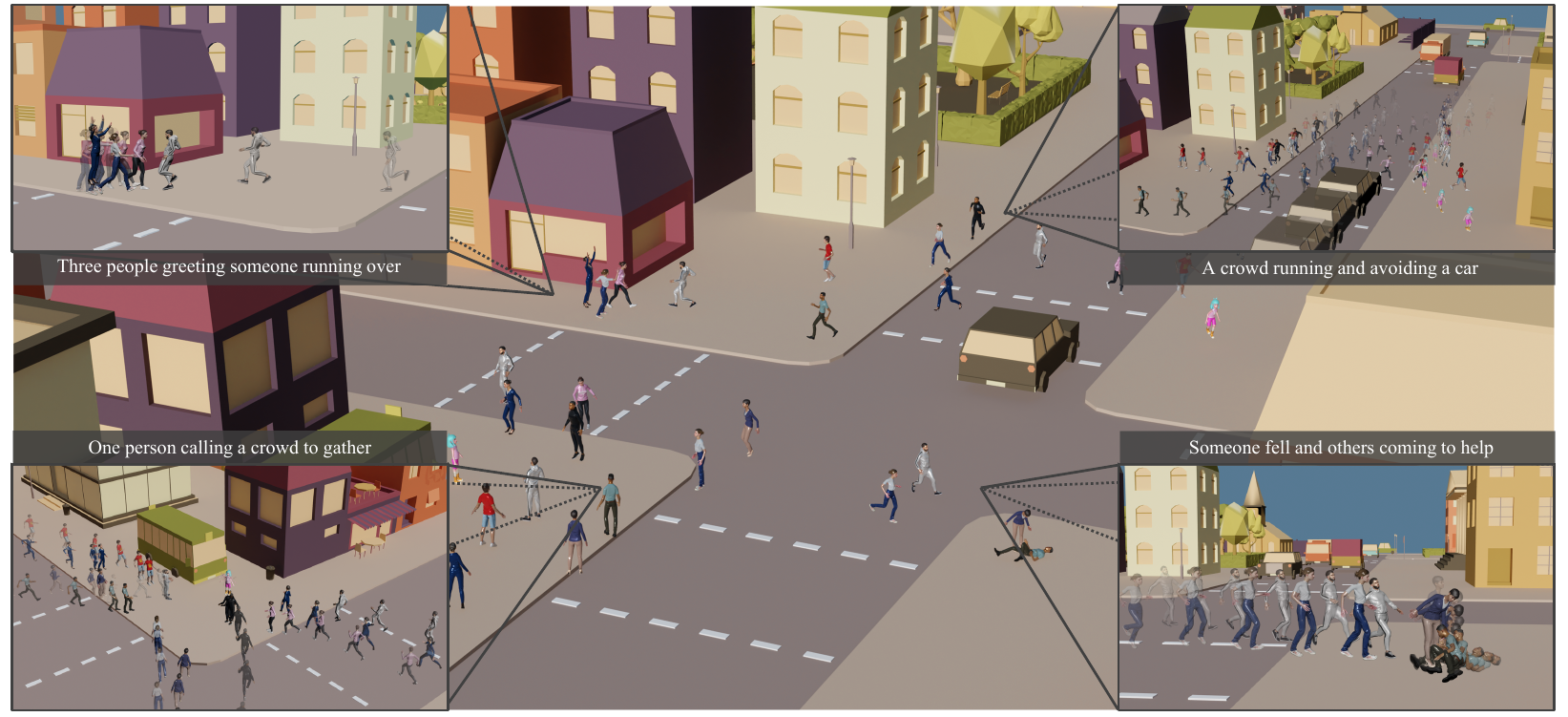}
 \caption{\textbf{\OMO} is a zero-shot, text-driven framework that enables generalizable planning and generation of crowd motions. Given a scene context, we aim to generate realistic crowd motions that fit the scene settings. }
  \label{fig:teaser}
  \vspace{-1em}
\end{figure*}

Fueled by the collections of large datasets~\cite{jiang2023full, bhatnagar2022behave, li2023object} with extensive text-motion pairs, text-guided human motion generation~\cite{zhou2023emdm, zhang2024motiondiffuse, chen2023executing, zhang2023generating, jiang2024motiongpt, tevet2022human} has made significant strides, yielding realistic and high-quality results.
Current techniques have demonstrated promising capabilities across various tasks, including single-person motion generation, multi-person motion generation, human-object interaction, and human-scene interaction. 
However, many of these approaches treat all individuals as a single, unified entity, overlooking scenarios where multiple groups exist within a scene, each separately responding to the specific context of an event.
To address this research gap, we focus on the task of \textbf{collective motion generation}$-$which involves (i) \textit{organizing individuals into distinct groups}, (ii) \textit{positioning these groups within a scene}, (iii) \textit{assigning relevant activities to each group}, and (iv) \textit{generating the corresponding human motions based on text prompts.} An example of this task is illustrated in Fig.~\ref{fig:teaser}.

While training a model in a supervised manner might seem like an intuitive solution, it would require datasets specifically tailored for collective motion generation. 
Given the complexity of this task and the limited size of existing 3D motion datasets~\cite{jiang2023full, bhatnagar2022behave, li2023object}, constructing and annotating such a dataset appears impractical.
Even if such a dataset were feasible, enumerating all possible events to cover every scenario would be a daunting challenge, restricting the model to a set of predefined situations.

This raises a crucial question: \textbf{Can we effectively perform collective motion generation without relying on such a dataset?} 
The challenge lies in the difficulty of directly learning the correspondence between text prompts, scene planning, and motion generation.
Our key insight is that \textit{scene planning and motion generation can be decoupled.} 
Specifically, the grouping of individuals, their positioning within the scene, and the assignment of activities can be directly derived from the text prompts.
Meanwhile, the motion of each individual can be guided by their specific activities, which can be represented as trajectory paths, enabling a more flexible and scalable approach for collective motion generation.

Motivated by this insight, we propose \OM, a novel zero-shot, text-guided framework for collective motion generation. 
Our method introduces two key components:
\textbf{1) Crowd scene planner:} 
\yk{While scene planning, \ie, group division and activities assignment, cannot be achieved through supervised training, we observe that this information can be derived from large language models that are independent of 3D motion datasets. 
To this end, we first utilized GPT-4 to organize individuals into distinct groups and position them appropriately within the scene.
While GPT-4 provides valuable high-level guidance for scene planning, it lacks the essential low-level understanding of human motion.
To address this gap, we propose integrating an SMPL-based joint prior, which enables the generation of context-appropriate activities.
These activities are represented as a combination of joint trajectories and textual descriptions, which serve as spatial control signals for the subsequent motion generation process.}
\textbf{2) Collective motion generator:} 
To improve the controllability and realism of the generated motions, we introduce the collective motion generator. 
\yk{This component integrates the spatial control signals into a transformer-based network in a joint-wise manner.
It starts by injecting both text and joint controls into noisy motion sequences, then uses a multi-head attention mechanism for each joint to improve stability.
This design helps maintain the spatial constraints during the multi-step denoising process, giving precise control over the generated motion.}

We extensively evaluate \OMO across various collective human motion scenarios and compare its performance with existing methods. Our experimental results demonstrate that \OMO significantly outperforms previous approaches, effectively grouping individuals, and generating realistic, event-aligned motion sequences that adhere to spatial constraints. 
\section{Related Works}
\textbf{Human motion generation.}
Variational Autoencoders (VAEs) and diffusion models are widely used in motion generation, leveraging diverse control prompts to create high-quality human motions. Motion prediction~\cite{gopalakrishnan2019neural, sun2021action, guo2023back, barquero2023belfusion, ahn2023can, DiffPred:2023, diller2022charposes, yan2023gazemodiff, wang2023gcnext, mao2020history, barsoum2018hp, chen2023humanmac, jiang2023motiondiffuser, sun2023towards}, for example, generate complete sequences from incomplete historical data or partial movements, ensuring the generated motions are smooth and realistic. A range of external controls, such as action categories~\cite{guo2020action2motion, petrovich2021action, zhao2023modiff, kulal2022programmatic}, music~\cite{li2021ai,  zhuang2022music2dance, siyao2022bailando, okamura2023dance, yao2023dance}, text~\cite{petrovich2022temos, tevet2022motionclip, tevet2022human, kim2023flame, athanasiou2022teach, lou2023diversemotion, zhang2024finemogen, shi2023generating, lu2023humantomato, ren2024insactor, zhang2024motiondiffuse, jiang2024motiongpt, hoang2024motionmix, yazdian2023motionscript, petrovich2024multi, lin2022ohmg, zhang2023remodiffuse}, scenes~\cite{huang2023diffusion, lim2023mammos, liu2023revisit, xiao2023unified}, objects~\cite{diller2023cghoi, li2023controllable, tendulkar2023flex, hao2024hand, lin2023handdiffuse, pi2023hierarchical, peng2023hoi, shimada2023macs}, and trajectories~\cite{karunratanakul2023guided, rempeluo2023tracepace}, further allows for stylized and customized motions. 
Additionally, recent innovations have introduced unified motion models~\cite{zhang2024large} that combine multi-modal inputs and multitask learning to enhance the versatility and controllability of individual human motion generation.

Recently, multi-person motion generation has gained increased attention. Some methods, like unsupervised motion completion for groups~\cite{NEURIPS2023_f4b52b45}, fail to capture the semantic and spatial dynamics of interactions due to reliance on limited-themed datasets lacking comprehensive annotations. Moreover, while two-person text-motion datasets such as DLP~\cite{cai2023digital} and InterHuman~\cite{liang2023intergen} have been introduced, methods depending on them effectively model pairwise interactions but struggle to extend to larger groups, failing to address broader inter-human dynamics. InterControl~\cite{wang2024intercontrol} makes initial strides to create multi-person interactions by explicitly controlling joint positions as per predefined motion plans, but its effectiveness diminishes in scenarios involving more than two people or larger crowds due to requiring heavy manual work. 
Consequently, existing methods expose significant shortcomings in realistically generating crowd motions, highlighting the need for more adaptable and logical motion planning, as well as efficient motion controlling and generation strategies.

\noindent\textbf{Controllable motion generation.}
\emph{Crowd Motion Generation} requires precise management of crowd dynamics and individual motion semantics, with the integration of spatial constraints into text-driven motion generation model posing significant challenges. Successful methods need to align with textual descriptions, adhere to spatial controls, maintain natural motion, and respect human body prior. Existing methods like MDM~\cite{tevet2022human} and PriorMDM~\cite{shafir2023human}, which rely on inpainting techniques, predict missing motion components from observed data but often fail to effectively manipulate joints other than the pelvis or handle sparse spatial constraints. GMD~\cite{karunratanakul2023guided} addresses the issue of sparse guidance with dense guidance propagation technique yet struggles to flexibly control spatial constraints across various joints and frames. OmniControl~\cite{xie2023omnicontrol} and InterControl~\cite{wang2024intercontrol} utilize a hybrid method combining classifier guidance with a ControlNet~\cite{zhang2023adding}, enhancing spatial signal integration and control accuracy. However, this approach impacts the realism of the motion. Inspired by these insights, we designed a transformer-based diffusion model that incorporates efficient control signal processing, a customized attention mechanism, and robust joint loss propagation. This configuration achieves an optimal balance between high-quality motion generation and precise model control capabilities.

\section{Methodology}

\begin{figure*}[t]
  \centering
  \includegraphics[width=\linewidth]{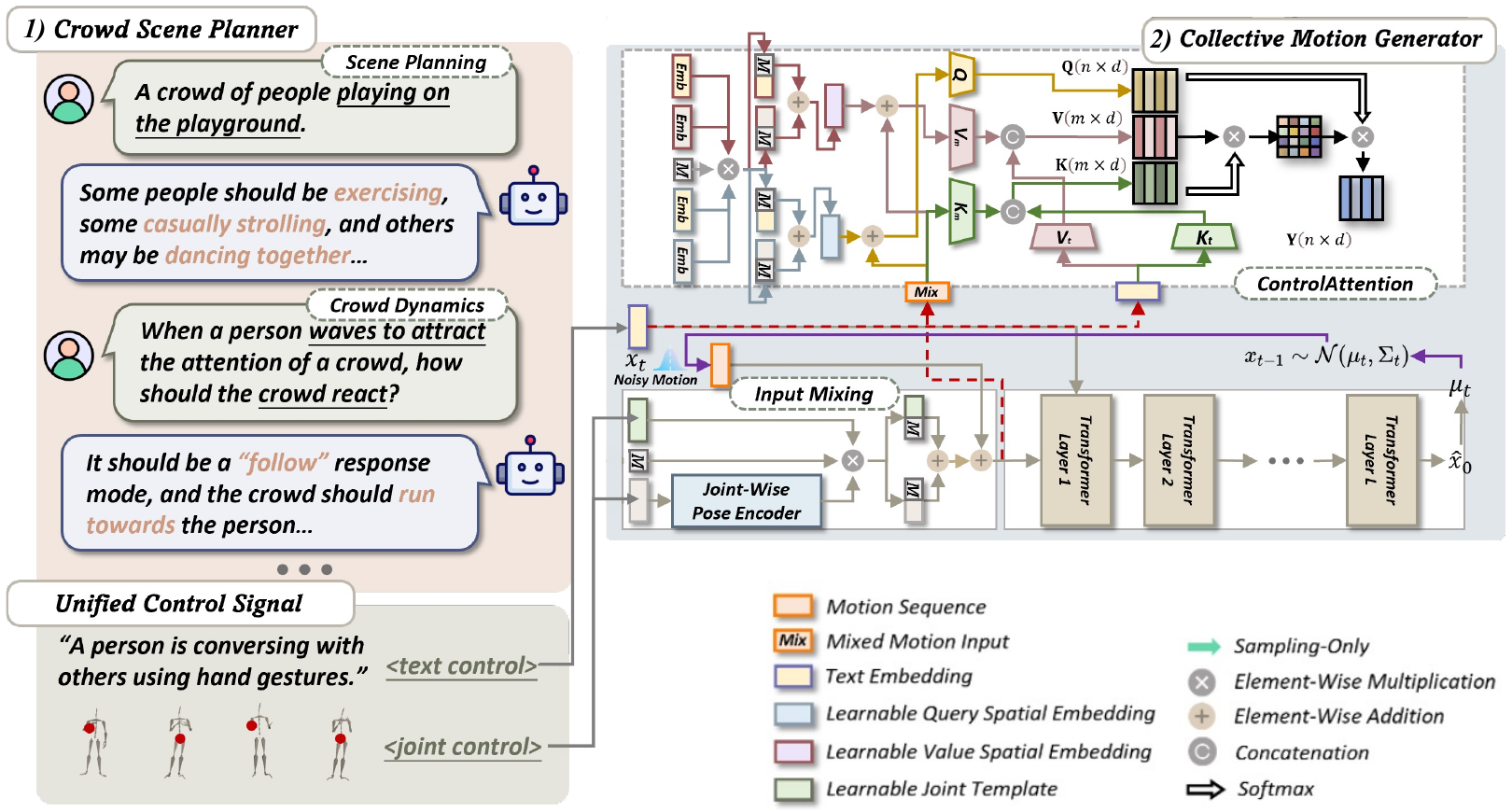}
  \vspace{-5 mm}
  \caption{\textbf{Overview of \OM.} The \textbf{\OMO} framework comprises two main components: \textbf{1) Crowd Scene Planner}, which uses a Large Language Model (LLM) to interpret and arrange crowd motions based on textual requirements from the user. This component then provides unified control signals in both textual and spatial formats. \textbf{2) Collective Motion Generator}, which leverages these control signals to manipulate and generate realistic individual motions.}
  \label{fig:pipeline}
  \vspace{-6 mm}
\end{figure*}

Give two textual prompts, $y_{\text{scene}}$ and $y_{\text{event}}$, which describe a scene and a perturbation event respectively, along with a specified total number of individuals $n$, our goal is to generate the motion sequences of these individuals as they interact within groups when the event occurs. 
To accomplish this, we introduce \textit{\textbf{\OM}}, a zero-shot, text-driven framework that naturally divides people into distinct groups and generates their collective motion sequences based on the texts. The overview of \OMO is illustrated in Fig.~\ref{fig:pipeline}. 
In the following sections, we first present the foundational concepts that underpin our method in Sec.~\ref{sec:preliminaries}. 
Next, we discuss the key components, including (1) the crowd scene planner (Sec.~\ref{sec:planner}), which uses a large language model (LLM) to transform the information derived from the textual prompt into unified spatial control signals, and (2) the collective motion generator (Sec.~\ref{sec:generator}), which utilizes these control signals to produce controllable human motion sequences. 
Finally, we discuss the training objectives in Sec.~\ref{sec:objectives}

\subsection{Preliminaries}
\label{sec:preliminaries}

\textbf{Diffusion process for motion generation.}
Diffusion models, which are a class of generative models, have shown great success in generating high-quality images from textual descriptions. 
Recent work has extended this approach to more complex tasks, such as generating human motion sequences.
Specifically, \cite{tevet2022human} proposes using diffusion models to generate entire motion sequences by predicting all of the poses at once.
In these models, the idea is to learn how to reverse a diffusion process, where noise is gradually added to a clean data sample, so that we can start with pure noise and "denoise" it step by step until we get back to a clean sample.

Mathematically, at each step $t$ of the diffusion process, the model predicts the distribution of the next step (moving toward the clean sample):
\begin{equation}
     P_{\theta}(\vx_{t-1}| \vx_{t}, \vp) = \mathcal{N}(\vmu_t(\theta), (1-\alpha_{t})\mI),
\end{equation}
where $\vx_t$ represents the noisy motion at step $t$, $\alpha_{t}$ is a parameter that adjusts the amount of noise at each step, and $\vmu_t(\theta)$ is the predicted mean of the distribution for the next step.

\begin{figure}[t]
  \centering
  \includegraphics[width=\linewidth, keepaspectratio]{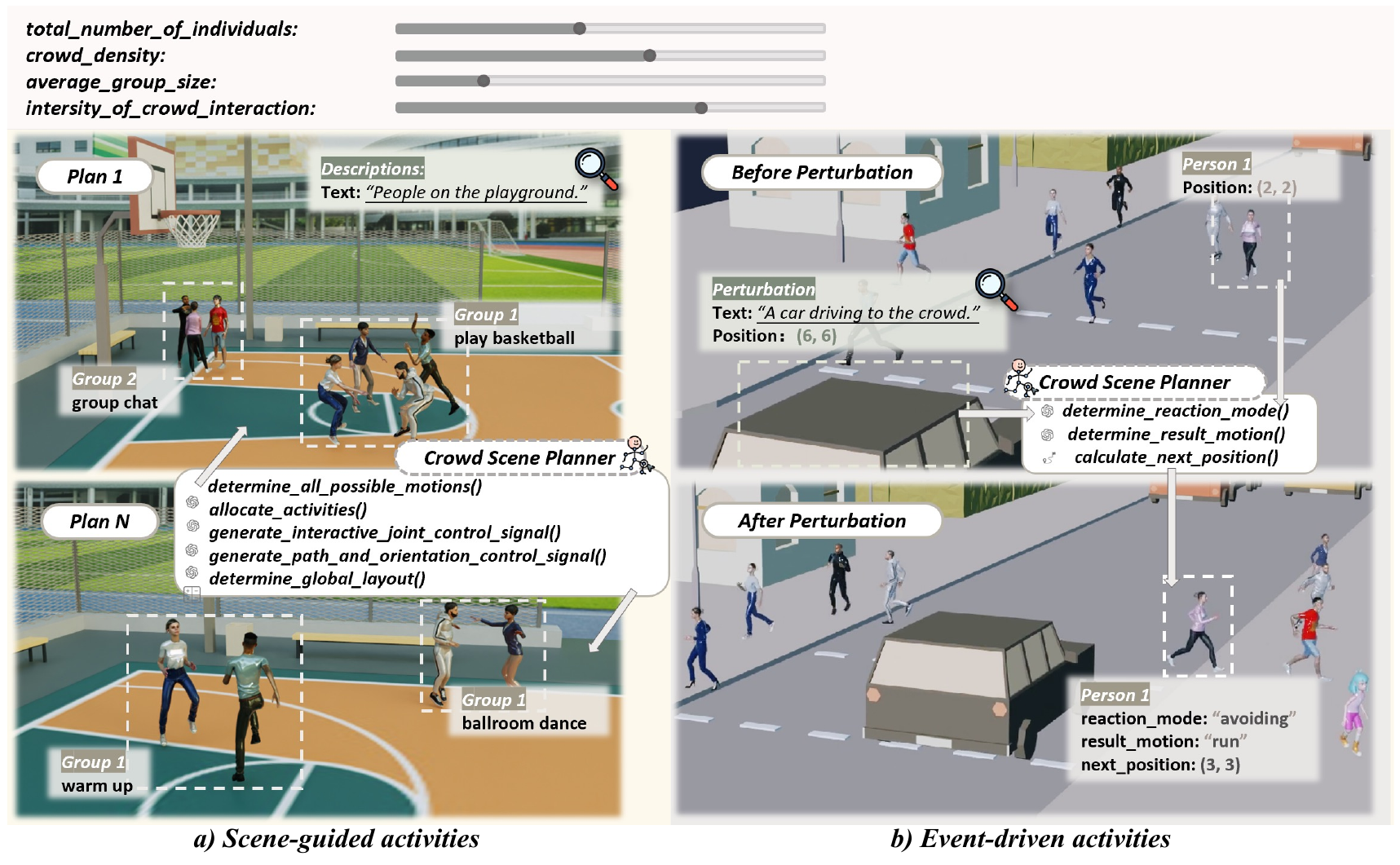}
  \vspace{-1.5em}
  \caption{\textbf{Scene-guided activities and event-driven activities.} The \textbf{Crowd Scene Planner} is able to deal with crowd motions effectively at both the scene and motion levels. It manages both scene-guided and event-driven activities, ensuring realistic and coherent crowd scenarios. Best viewed in PDF with zoom-in.}
  \label{fig:activities}
  \vspace{-1.5em}
\end{figure}
\subsection{Crowd Scene Planner}
\label{sec:planner}

Existing human motion generation pipelines, whether for single or multiple people~\cite{wang2024intercontrol, zhang2024motiondiffuse, chen2023executing, zhang2023generating}, typically assume that all individuals are grouped together as a single unit. This assumption makes it challenging to generate diverse behaviors for different groups, especially in scenarios where different groups need to act in unique ways based on the event—something that is common in real-world applications and essential for effective collective motion generation. Consequently, two key aspects need to be addressed to improve from these pipelines: (1) semantically dividing individuals into separate groups while maintaining their interactions, and (2) accurately assigning context-appropriate activities (e.g., dancing, exercising, or walking) to each group, while also ensuring that their spatial relationships, movement trajectory, and interactions are realistically modeled. In our method, we leverage the extensive crowd motion knowledge embedded in GPT-4~\cite{achiam2023gpt} to tackle these challenges. 

\noindent\textbf{Semantical group division.}
We begin by organizing the individuals into distinct groups before generating their motion sequences.
To do so, we define several hyper-parameters to determine the rough organization of the crowd:
\begin{equation}
    \mathcal{C}_{\text{params}} = \{n, s, \sigma, \alpha\},
\end{equation}
where $n$ is the total number of individuals; $s$ denotes the average group size, which indicates whether the model tends to form larger groups or smaller groups; $\sigma$ is the crowd density, determining whether more or fewer groups are generated; $\alpha$ represents the intensity of crowd interaction, which measures the level of physical interaction within each group. 
\yk{It is important to note that $s$ and $\sigma$ are correlated but not dependence. For instance, both $s$ and $\sigma$ can be large if several groups are large while the others are very small.}
Examples are provided in the upper section of Fig.~\ref{fig:activities}.
Users can manually set these parameters or choose to use our crowd scene planner, which will automatically adjust them if not provided by the user. 

\noindent\textbf{Context-appropriate activities assignment.}
\yk{After separating individuals into groups}, we now generate detailed activities $\mathcal{E}_{\text{plan}} = (\mathcal{E}^{s}, \mathcal{E}^{t})$ for each group based on the input textual prompts $y_{\text{scene}}$ and $y_{\text{event}}$. As shown in Fig.~\ref{fig:pipeline}, the spatial control signals $\mathcal{E}^{s} \in \mathbb{R}^{n \times f \times J \times 3}$ define the joint and trajectory plans for each person, where $f$ represents the number of motion frames and $J$ is the number of joints. We design $\mathcal{E}^{s}$ in the 3D space to provide more accurate spatial information compared to 2D images. On the other hand, $\mathcal{E}^{t}$ contains the corresponding textual descriptions of the motions. 
Specifically, we categorize these activities $\mathcal{E}_{\text{plan}}$ into two types: scene-guided activities and event-driven activities.

\noindent\textit{(1) Scene-guided activities.}
Based on the scene description $y_{\text{scene}}$, our crowd scene planner initially uses a large language model, like GPT-4, to generate \yk{multiple} possible motion plans, denoted as $\mathcal{M}$. As illustrated in Fig.~\ref{fig:activities}(a), for \yk{each motion plan in $\mathcal{M}$, we then (i) assign the activities, (ii) generate the interactive joint control signals, and (iii) define the global layout for each group via GPT-4.}


\noindent\textit{(2) Event-driven activities.}
Once the scene-guided activities are established, our next objective is to model how individuals will behave when a perturbation, as described in $y_{\text{event}}$, occurs. To achieve this, we identify six primary response patterns\footnote{\yk{While real-world scenarios are more complex than our six event patterns, we find that breaking them into smaller events allows our patterns to cover most situations.}} based on different types of events:
\textbf{1)} \emph{Following}, where the crowd mimics the actions of a leader; \textbf{2)} \emph{Avoiding}, where the crowd moves to dodge a fast-moving obstacle, such as a vehicle; \textbf{3)} \emph{Queuing}, where the crowd forms an organized, orderly line; \textbf{4)} \emph{Encircling}, where the crowd gathers around an interesting object or event; \textbf{5)} \emph{Passing}, where the crowd makes temporary adjustments to bypass a disturbance and then resumes their original activities; \textbf{6)} \emph{Random}, where there is no clear, coordinated movement within the crowd. 
Leveraging the capabilities of GPT-4, along with the embedded crowd knowledge derived from the scene-guided activities, our planner selects the most appropriate response pattern based on $y_{\text{event}}$. It then updates the motions of the affected individuals and applies trajectory modification algorithms to adjust their positions according to the selected response pattern, as shown in Fig.~\ref{fig:activities}(b). \yk{It's important to note that our event patterns are flexible. Users can modify or add patterns based on their application needs.}

\begin{table*}[t]
\centering
\caption{\textbf{Controllable motion generation quantitative results on HumanML3D test set. }`$\uparrow$'(`$\downarrow$') indicates that the values are better if the metric is larger (smaller). `$\rightarrow$' means closer to real data is better. The best result are in \textbf{bold}.}
\vspace{-1em}
\label{tab:control}
\resizebox{\textwidth}{!}{%
\begin{tabular}{c|c|ccccccc}
\toprule
\multirow{2}{*}{\centering\textbf{Method}} & \multirow{2}{*}{\centering\textbf{Joint}} & \multirow{2}{*}{\centering\textbf{FID $\downarrow$}} & \textbf{R-precision} & \multirow{2}{*}{\centering\textbf{Diversity $\rightarrow$}} & \multirow{2}{*}{\centering\textbf{Foot skating ratio} $\downarrow$} & \textbf{Traj. err.} & \textbf{Loc. err.} & \textbf{Avg. err.} \\ 
 &  &  & \textbf{(Top-3) $\uparrow$} &  &  & \textbf{(50 cm) $\downarrow$} & \textbf{(50 cm) $\downarrow$} & \textbf{(m) $\downarrow$} \\
\midrule
Real &  & 0.002 & 0.797 & 9.503 & 0.000 & 0.000 & 0.000 & 0.000 \\ 
\midrule
PriorMDM~\cite{shafir2023human} & \multirow{4}{*}{\centering Pelvis} & 0.475 & 0.583 & 9.156 & 0.0897 & 0.3457 & 0.2132 & 0.4417 \\
GMD~\cite{karunratanakul2023guided} & & 0.576 & 0.665 & 9.206 & 0.109 & 0.0931 & 0.0321 & 0.1439  \\
OmniControl~\cite{xie2023omnicontrol} & & 0.218 & 0.687 & 9.422 & \textbf{0.0547} & 0.0387 & 0.0096 & 0.0338  \\
InterControl~\cite{wang2024intercontrol} & & 0.159 & 0.671 & \textbf{9.482} & 0.0729 & 0.0132 & 0.0004 & 0.0496 \\
\emph{\textbf{CrowdMoGen}} & & \textbf{0.132} & \textbf{0.784} & 9.109 & 0.0762 & \textbf{0.0000} & \textbf{0.0000} & \textbf{0.0196} \\  
\midrule
OmniControl~\cite{xie2023omnicontrol} & \multirow{3}{*}{\centering Random One} & 0.310 & 0.693 & \textbf{9.502} & \textbf{0.0608} & 0.0617 & 0.0107 & 0.0404  \\
InterControl~\cite{wang2024intercontrol} & & 0.178 & 0.669 & 9.498 & 0.0968 & 0.0403 & 0.0031 & 0.0741 \\
\emph{\textbf{CrowdMoGen}} & & \textbf{0.147} & \textbf{0.781} & 9.461 & 0.0829 & \textbf{0.0000} & \textbf{0.0000} & \textbf{0.0028} \\  
\midrule
InterControl~\cite{wang2024intercontrol} & \multirow{2}{*}{\centering Random Two} & 0.184 & 0.670 & \textbf{9.410} & 0.0948 & 0.0475 & 0.0030 & 0.0911 \\
\emph{\textbf{CrowdMoGen}} & & \textbf{0.178} & \textbf{0.777} & 9.149 & \textbf{0.0865}  & \textbf{0.0000} & \textbf{0.0000} & \textbf{0.0027} \\  
\midrule
InterControl~\cite{wang2024intercontrol} & \multirow{2}{*}{\centering Random Three} & 0.199 & 0.673 & \textbf{9.352} & 0.0930 & 0.0487 & 0.0026 & 0.0969 \\
\emph{\textbf{CrowdMoGen}} & & \textbf{0.192} & \textbf{0.778} & 9.169 & \textbf{0.0871} & \textbf{0.0000} & \textbf{0.0000} & \textbf{0.0030} \\  
\bottomrule
\end{tabular}%
}
\vspace{-1em}
\end{table*}

\subsection{Collective Motion Generator}
\label{sec:generator}

After generating context-appropriate activities $\mathcal{E}_{\text{plan}} = (\mathcal{E}^{s}, \mathcal{E}^{t})$, applying them to produce human motions presents an additional challenge. Previous works~\cite{wang2024intercontrol, xie2023omnicontrol, shafir2023human} have demonstrated promising spatial controllability using ControlNet~\cite{zhang2023adding}; however, ControlNet does not account for 3D spatial properties, which limits its effectiveness in controlling 3D/4D human motion generation. To overcome these limitations, we propose to integrate 3D spatial control signals $\mathcal{E}_{\text{plan}}$ into the transformer network to improve the quality of generated human motions.

\noindent\textbf{InputMixing.} 
We follow recent methods~\cite{tevet2022human, zhang2024finemogen,ho2020denoising} to employ diffusion model~\cite{ho2020denoising} and transformer as our backbone to recover motion sequence $\mathcal{T}(\mathbf{x}_{t}, t, \mathcal{E}_{\text{plan}} = \{\mathcal{E}^{s}, \mathcal{E}^{t}\})$ from a noisy input sequence $\mathbf{x}_{t}$, where $t$ denotes the time step. Nevertheless, this network by itself is insufficient to provide fine-grained control over the generated human motions, as it lacks the necessary conditioning information. To address this gap, we propose injecting the spatial control signal $\mathcal{E}^{\text{plan}}$ into the noisy sequence $\mathbf{x}_{t}$. \yk{Specifically, we introduce two joint-wise encoder, $E^{s}$ and $E^{\textbf{x}}$, to incorporate this additional conditioning knowledge:}
\begin{equation}
\label{eq:pre-inject}
    \mathbf{x}_{t,i,j} := \cdot E^{s}_{j}(\mathcal{E}^{s}_{i, j}) + E^{\textbf{x}}_{j}((\mathbf{x}_t)_{i,j}),
\end{equation}
where $i$ represents the $i$-th individual, $j$ denotes the $j$-th joint, 
and $\mathbf{x}_{t,i,j}$ represents the noisy sequence for $i$-th individual's $j$-th joint. 

\yk{Unfortunately, there's one issue: we've noticed that our GPT-based scene planner doesn't always generate trajectories for all the pre-defined joints, due to the limitations of the original GPT model. This creates an imbalance in control between joints with generated trajectories and those without. To fix this, we classify joints with generated trajectories as active (controlled) and the rest as inactive (uncontrolled), using a binary mask $M \in \{0, 1\}^{n \times f \times J}$ to represent this. We then redesign the noise injection as described in Eq.~\ref{eq:pre-inject} to:}
\begin{equation}
    \mathbf{x}_{t,i,j} := M_{i, j}\cdot E^{s}_{j}(\mathcal{E}^{s}_{i, j}) + (1 - M_{i, j})\cdot c_{i, j}^{\text{temp}} + E^{\textbf{x}}_{j}((\mathbf{x}_t)_{i,j}),
\end{equation}
where $c^{\text{temp}} \in \mathbb{R}^{f \times J \times L}$ denotes a learnable joint template, balancing the controlled and uncontrolled joints. $L$ is the dimension of the latent feature for each joint.

\noindent\textbf{ControlAttention.}
While our proposed \textit{InputMixing} technique successfully injects 3D awareness into the denoising process, the network may still struggle with stability due to the non-linear nature of the multi-step denoising procedure. As a result, the model tends to generate unstable trajectories, as shown in Tab.~\ref{tab:ablation}. To address this challenge, we introduce a novel attention mechanism called \textit{ControlAttention}, based on Efficient Attention~\cite{shen2021efficient,zhang2024motiondiffuse}. Specifically, we design $J$ attention heads, each corresponding to a different joint, enabling joint-wise modeling. \yk{The reasons for this design are two-fold: (1) it lets us differentiate the attention between active and inactive joints, and (2) it enables each attention head to focus more effectively on the relevant joint, making better use of the corresponding joint-specific control signal.} 
\begin{equation}
    \begin{aligned}
    \text{emb}_{Q} &= \text{emb}_{Q}^{\text{mask}}M + \text{emb}_{Q}^{\text{control}}(1-M), \\
    \text{emb}_{V} &= \text{emb}_{V}^{\text{mask}}M + \text{emb}_{V}^{\text{control}}(1-M),
    \end{aligned}
\end{equation}
where $\text{emb}_{Q}^{\text{mask}}, \text{emb}_{V}^{\text{mask}} \in \mathbb{R}^{f \times J \times L}$ are designed to capture and integrate spatial information from inactive joints, whereas $\text{emb}_{Q}^{\text{control}}$ and $\text{emb}_{V}^{\text{control}}$ focus specifically on the active joints. In this way, the combined embeddings $\text{emb}_{Q}$ and $\text{emb}_{V}$ encapsulate comprehensive spatial information for both active and inactive joints. These embeddings ($\text{emb}_{Q}$, $\text{emb}_{V}$) are then added to the module input, where they are used to compute the $Q$ and $V$ vectors.

\begin{table*}[t]
  \begin{minipage}[t]{0.49\linewidth}
\caption{\textbf{Text-to-motion evaluation on KIL-ML test set.} `$\uparrow$'(`$\downarrow$') indicates that the values are better if the metric is larger (smaller). `$\rightarrow$' means closer to real data is better. The best results are in \textbf{bold}.}
\vspace{-1em}
\label{tab:kitml_text}
\resizebox{0.9\textwidth}{!}{

\begin{tabular}{@{}c|ccc@{}}
\toprule
\multirow{2}{*}{\centering\textbf{Method}} & \multirow{2}{*}{\centering\textbf{FID $\downarrow$}} & \textbf{R-precision} & \multirow{2}{*}{\centering\textbf{Diversity $\rightarrow$}} \\ 
 &  & \textbf{(Top-3) $\uparrow$} & \\
\midrule
Real & 0.031 & 0.779 & 11.08 \\
\midrule
T2M \cite{guo2022generating} & 3.022 & 0.681 & 10.72 \\
MotionDiffuse \cite{zhang2024motiondiffuse} & 1.954 & 0.739 & \textbf{11.10} \\
MLD \cite{chen2023executing} & 0.404 & 0.734 & 10.80 \\
T2M-GPT \cite{zhang2023generating} & 0.514 & 0.745 & 10.92 \\
MotionGPT \cite{jiang2024motiongpt} & 0.510 & 0.680 & 10.35 \\
MDM \cite{tevet2022human} & 0.497 & 0.396 & 10.84 \\
\midrule
PriorMDM \cite{shafir2023human} & 0.830 & 0.397 & 10.54 \\
GMD \cite{karunratanakul2023guided} & 1.537 & 0.385 & 9.78 \\
OmniControl \cite{xie2023omnicontrol} & 0.702 & 0.397 & 10.93 \\
InterControl \cite{wang2024intercontrol} & 0.580 & 0.397 & 10.88 \\ 
\emph{\textbf{CrowdMoGen}} & \textbf{0.217} & \textbf{0.777} & 10.33 \\ 
\bottomrule
\end{tabular}
\vspace{-20pt}
}
    \vspace{-1em}
  \end{minipage}
  ~
  \begin{minipage}[t]{0.49\linewidth}
\centering
\caption{\textbf{Text-to-motion evaluation on HumanML3D test set.}}
\label{tab:humanml3d_text}
\resizebox{0.9\textwidth}{!}{
\begin{tabular}{@{}c|ccc@{}}
\toprule
\multirow{2}{*}{\centering\textbf{Method}} & \multirow{2}{*}{\centering\textbf{FID $\downarrow$}} & \textbf{R-precision} & \multirow{2}{*}{\centering\textbf{Diversity $\rightarrow$}} \\ 
 &  & \textbf{(Top-3) $\uparrow$} & \\
\midrule
Real & 0.002 & 0.797 & 9.503 \\
\midrule
T2M \cite{guo2022generating} & 1.067 & 0.740 & 9.188 \\
MotionDiffuse \cite{zhang2024motiondiffuse} & 0.630 & 0.782 & 9.410 \\
MLD \cite{chen2023executing}  & 0.473 & 0.772 & 9.724 \\
PhysDiff \cite{yuan2023physdiff}  & 0.413 & 0.631 & - \\
T2M-GPT \cite{zhang2023generating} & 0.116 & 0.775 & 9.761 \\
MotionGPT \cite{jiang2024motiongpt} & 0.232 & 0.778 & 9.528 \\
MDM \cite{tevet2022human} & 0.544 & 0.611 & 9.446 \\
\midrule
PriorMDM \cite{shafir2023human} & 0.540 & 0.640 & 9.160 \\
GMD \cite{karunratanakul2023guided}& 0.212 & 0.670 & 9.440 \\
OmniControl \cite{xie2023omnicontrol} & 0.218 & 0.687 & 9.422 \\
InterControl \cite{wang2024intercontrol} & 0.159 & 0.671 & \textbf{9.482} \\
\emph{\textbf{CrowdMoGen}} & \textbf{0.112} & \textbf{0.784} & 9.109 \\ \bottomrule
\end{tabular}
}
    \vspace{-1em}
  \end{minipage}
\end{table*}

\subsection{Training Objectives}
\label{sec:objectives}

To enhance motion precision and the naturalness of the animations, we extend the commonly used \emph{Whole Body Loss} $\mathcal{L}_\text{whole}$ by incorporating two additional loss components: \emph{Controlled Keypoint Loss} $\mathcal{L}_\text{con}$ and \emph{Foot Skating Loss} $\mathcal{L}_\text{foot}$.
Specifically, $\mathcal{L}_\text{whole}$ can be formulated as:
\begin{equation}
    \mathcal{L}_\text{whole} = \mathbb{E}[||\mathcal{T}(\mathbf{x}_{t}, t, \mathcal{E}_{\text{plan}}) - \mathbf{x}^{\text{gt}}||^2_2],
\end{equation}
where $\mathbf{x}^{\text{gt}}$ denotes the ground-truth motions.

\noindent\textit{(1) Controlled Keypoint Loss} It computes the 3-dimensional Euclidean distance between controlled keypoints and their corresponding global signals:
\begin{equation}
    \mathcal{L}_\text{con} = |M| \cdot \|\mathcal{E}^{s} - \hat{\mathbf{x}} ^{g} \|_{2},  \quad |M| = \sum_{i=0, ..., n}M_i
\end{equation}
where $\hat{\mathbf{x}}^g \in \mathbb{R}^{n \times f \times J \times 3}$ is the predicted output.

\begin{figure*}[t]
  \centering
  \includegraphics[width=\linewidth, keepaspectratio]{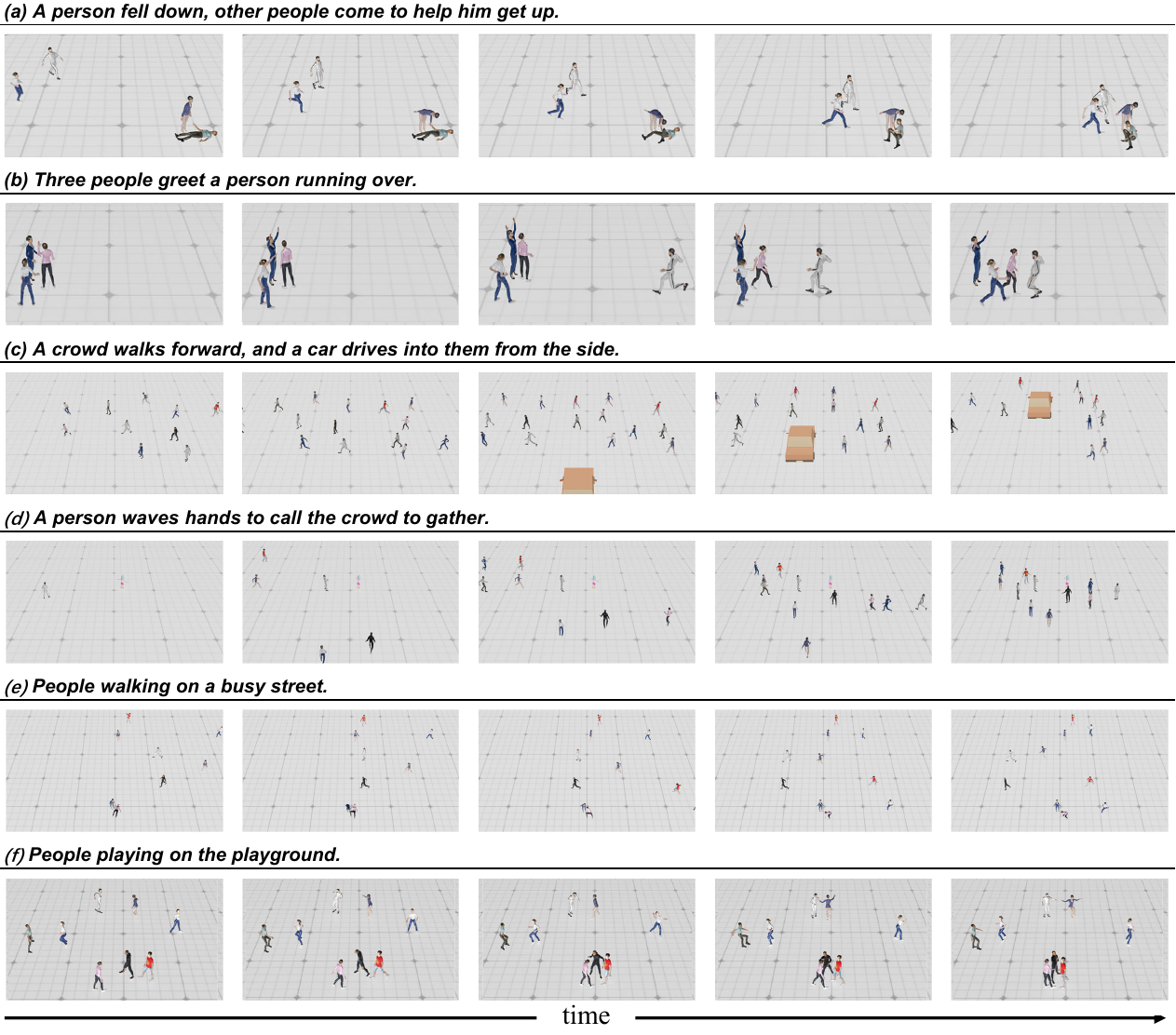}
  \vspace{-2em}
  \caption{\textbf{Qualitative Visualizations.} Displayed are selected frames from the crowd motion sequences generated by our proposed \textbf{\OM}. It effectively creates scenarios involving multi-person close interactions \emph{(a)-(b)}, dynamic crowd movements \emph{(c)-(d)}, and complex crowd scenes \emph{(e)-(f)} that accurately and naturally reflect the specified scene descriptions.}
  \label{fig:qualitative}
  \vspace{-3mm}
\end{figure*}

\noindent\textit{(2) Foot Skating Loss} It assesses the sliding of the foot joint, supervising the 3-dimensional Euclidean distance between the foot joints in two consecutive frames if their heights are below a certain threshold $h_{\text{thresh}}$:
\begin{equation}
\small
\begin{split}
    \mathcal{L}_\text{foot} &= \sum_{i=1}^{f-1}  \\
 &\left( \sum_{j \in \{j_{\text{left}}, j_{\text{right}}\}} \left((\hat{\mathbf{x}}^g)_{i, j, 2} < h_{\text{thresh}}\right) \cdot \left\| (\hat{\mathbf{x}}^g)_{i+1, j} - (\hat{\mathbf{x}}^g)_{i, j} \right\|_2 \right),
    \end{split}
\end{equation}
where the indices $j_{\text{left}}$ and $j_{\text{right}}$ denote the left and right foot keypoints, respectively.

The overall training objectives can be formulated as:
\begin{equation}
\mathcal{L}=\lambda_\text{whole} \cdot \mathcal{L}_\text{whole} + \lambda_\text{con} \cdot \mathcal{L}_\text{con} + \lambda_\text{foot} \cdot \mathcal{L}_\text{foot},
\end{equation}
where $\lambda_\text{whole},\lambda_\text{con},\lambda_\text{foot}$ are hyper-parameters. 
Note that our training objectives only apply to the collective motion generator while the crowd scene planner is a training-free stage via pre-trained large language models.

\noindent\textbf{Inference Strategies.} During the inference stage, we apply a 50-step denoising process to synthesize motion sequences while using classifier-free guidance for better generation quality. In addition, we use (Inverse Kinematics) IK guidance~\cite{wang2024intercontrol} in the last iterations for better spatial alignment.
\section{Experiments}

We now evaluate the efficiency of our \OMO across different scenarios and perform a comparative analysis with state-of-the-art text-to-motion pipelines.

\subsection{Datasets}

We apply HumanML3D~\cite{guo2022generating} and KIT Motion Language (KIT-ML)~\cite{plappert2016kit} datasets to evaluate the generation and control capabilities of our method and compare with existing SOTAs. Specifically, the HumanML3D dataset is a reannotated combination of the HumanAct12\cite{guo2020action2motion} and AMASS~\cite{mahmood2019amass} datasets, containing 14,616 motion sequences and 44,970 textual descriptions. The KIT-ML dataset consists of 3,911 motion sequences, paired with 6,363 natural language descriptions.


\begin{table*}[t]
\centering
\caption{\textbf{Ablation studies on the HumanML3D dataset.} `$\uparrow$'(`$\downarrow$') indicates that the values are better if the metric is larger (smaller). `$\rightarrow$' means closer to real data is better. The best results are in \textbf{bold}.}
\label{tab:ablation}
\resizebox{\textwidth}{!}{%
\begin{tabular}{c|c|ccccccc}
\toprule
\multirow{2}{*}{\centering\textbf{Item}} & \multirow{2}{*}{\centering\textbf{Method}} & \multirow{2}{*}{\centering\textbf{FID $\downarrow$}} & \textbf{R-precision} & \multirow{2}{*}{\centering\textbf{Diversity $\rightarrow$}} & \multirow{2}{*}{\centering\textbf{Foot skating ratio} $\downarrow$} & \textbf{Traj. err.} & \textbf{Loc. err.} & \textbf{Avg. err.} \\ 
 &  &  & \textbf{(Top-3) $\uparrow$} &  &  & \textbf{(20 cm) $\downarrow$} & \textbf{(20 cm) $\downarrow$} & \textbf{(m) $\downarrow$} \\
\midrule
(1) & \textbf{CrowdMoGen} & \textbf{0.127} & \textbf{0.784} & 9.109 & \uline{0.0762} & \textbf{0.0051} & \textbf{0.0000} & 0.0196 \\ 
\midrule
(2) & w/o ControlAttention & 0.150 & 0.775 & \textbf{9.446} & 0.1037 & 0.0069 & 0.0001 & 0.0223 \\ 
(3) & w/o InputMixing & 0.207 & 0.771 & 9.090 & 0.0919 & 0.0159 & 0.0003 & 0.0246 \\ 
(4) & w/o $\mathcal{L}_{\text{con}}$ & 0.219 & 0.761 & 9.143 & 0.1316 & 0.0174 & 0.0004 & 0.0254  \\ 
(5) & w/o $\mathcal{L}_{\text{foot}}$ & 0.131 & 0.777 & 9.244 & 0.1947 & 0.0058 & 0.0001 & \textbf{0.0147} \\ 
(6) & w/o IK guidance & 0.130 & 0.782 & 9.320 & \textbf{0.0751} & 0.5710 & 0.2720 & 0.2387 \\ 
(7) & w/ only ControlNet & 0.225 & 0.763 & 9.814 & 0.0915 & 0.0210 & 0.0005 & 0.0427 \\ 
\bottomrule
\end{tabular}
}
\vspace{-10pt}
\end{table*}
\subsection{Evaluation metrics}

Following~\cite{guo2022generating}, we use \emph{Frechet Inception Distance (FID)}, \emph{R-Precision}, and \emph{Diversity} to respectively evaluate the quality of generated motions, the similarity to the textual descriptions, and generation variability.
Additionally, we adopt the \emph{Foot Skating Ratio}, \emph{Trajectory Error}, \emph{Location Error}, and \emph{Average Error} metrics from OmniControl~\cite{xie2023omnicontrol} to evaluate the control accuracy.

\yk{Specifically, (1) \emph{FID} reflects the generation quality by calculating the distance between features extracted from real and generated motion sequences; (2) \emph{R-Precision}: A pool is formed from the accurate description of a generated motion and 31 unrelated descriptions from the test dataset. Euclidean distances between the motion’s features and those of each description in the pool are calculated and ranked. Accuracy is then averaged at the top-1, top-2, and top-3 levels, with successful retrievals marked by the ground truth description appearing within the top-k selections; (3) \emph{Diversity} measures the variability and richness of the generated motion sequences; (4) \emph{Foot Skating Ratio} is a metric used to quantify the amount of sliding or unrealistic movements of feet in animations; (5) \emph{Trajectory Error} refers to the proportion of trajectories that are unsuccessful, which occurs when any keyframe's location error exceeds a set threshold; (6) \emph{Location Error} represents the ratio of keyframe locations that do not fall within a threshold distance; (7) \emph{Average Error} is computed as the mean distance between the positions of generated motions and the actual keyframe positions observed during the key motion steps.}
 
\subsection{Implementation details}

We construct our text encoder by combining a CLIP ViT-B/32 text encoder with two transformer encoder layers. Our diffusion model operates with $1000$ time steps, where the variances $\beta_t$ range linearly from $0.0001$ to $0.02$. The motion decoder is implemented as a 4-layer transformer. We conduct the training on eight NVIDIA Tesla V100 GPUs, with each GPU processing $64$ samples, resulting in a total batch size of $512$. The model is trained on the KIT-ML and HumanML3D datasets for $40,000$ and $100,000$ iterations, respectively. We use the Adam optimizer with an initial learning rate of $2e^{-4}$, which is reduced to $2e^{-5}$ during the final $20\%$ of the training iterations. For all experiments presented in this paper, we set the weights $\lambda_{whole},\lambda_{con}, \lambda_{foot}$ to $1.0$.

\subsection{Comparison methods}

Given that existing techniques are unable to effectively address collective motion generation, particularly in stages like scene planning, we first compare \OMO with PriorMDM~\cite{shafir2023human}, GMD~\cite{karunratanakul2023guided}, OmniControl~\cite{xie2023omnicontrol}, and InterControl~\cite{wang2024intercontrol} in terms of controllable motion generation quality. We also compare with T2M~\cite{guo2022generating}, MotionDiffuse~\cite{zhang2024motiondiffuse}, MLD~\cite{chen2023executing}, T2M-GPT~\cite{zhang2023generating}, MotionGPT~\cite{jiang2024motiongpt}, and MDM~\cite{tevet2022human} on the quality of text-to-motion generation for more comprehensive comparisons. 
\yk{To evaluate the performance of our proposed scene planner, we also compare it with the baseline design (original GPT-4).}

\subsection{Quantitative comparisons}

\noindent\textbf{Evaluations on motion quality.}
We first evaluate the quality of the generated motion and present the results in Tab.~\ref{tab:control}. The findings show that our generated motion sequences closely align with the given spatial conditions, as indicated by the low "Traj. err" and "loc. err" values. Along with the "R-precision" comparisons, these results highlight \OM's ability to produce realistic, text-aligned motion sequences that adhere to specific spatial constraints.
Additionally, we compare our results with those of existing text-to-motion methods, with the detailed comparisons available in Tab.~\ref{tab:kitml_text} and \ref{tab:humanml3d_text}. Notably, \OMO outperforms the state-of-the-art approaches, even though it was not specifically designed for text-to-motion generation.

\begin{figure}[t]
  \centering
  \includegraphics[width=\linewidth, keepaspectratio]{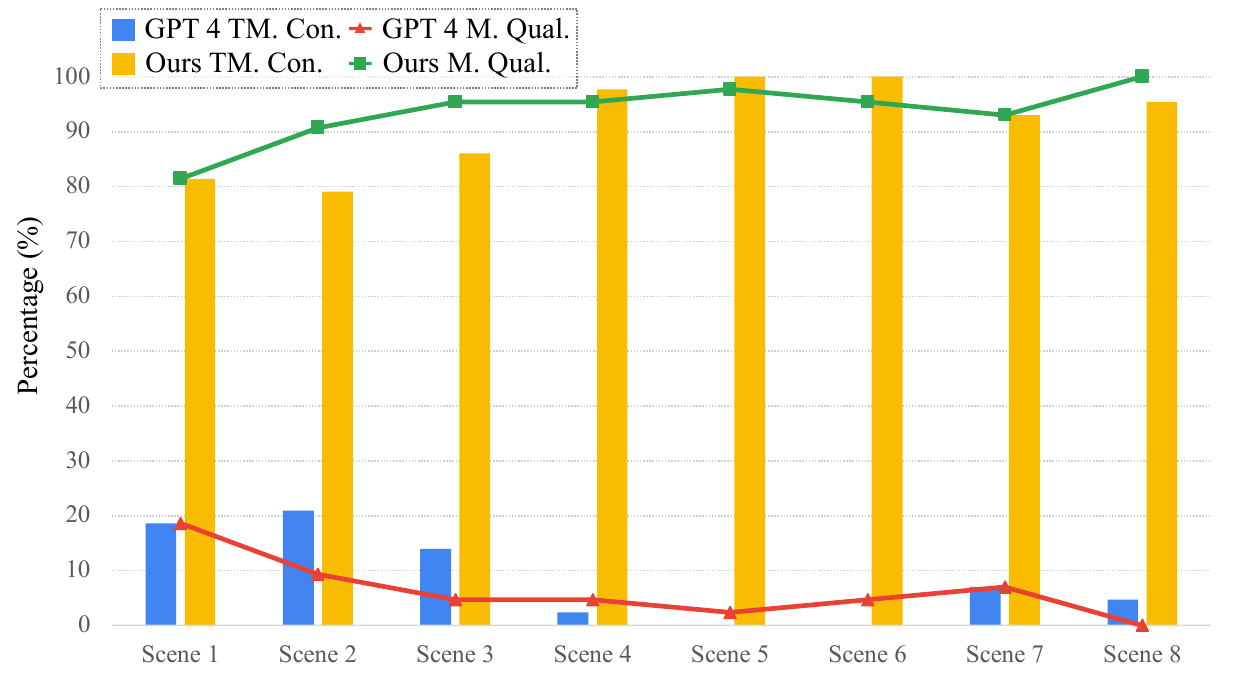}
  \caption{\textbf{User Study: Comparative Analysis of Planning Methods.} This chart presents the comparison results between our \textbf{Crowd Scene Planner} and plain GPT-4, based on participant preferences for text-motion consistency (TM. Con.) and motion quality (M. Qual.). The percentages reflect the proportion of participants who favored each method.}
  \vspace{-1em}
  \label{fig:userstudy}
\end{figure}
\noindent\textbf{User studies for scene planner.}
We also conducted user studies to assess the performance of our proposed crowd scene planner. In total, 50 volunteers participated, including animators, AI researchers, and gaming enthusiasts, aged between 20 and 35. Each participant was given textual descriptions of crowd scenes and asked to compare the crowd scene plans generated by our method with those produced by the standard GPT-4 model, evaluating them based on motion quality ("M. Qual.") and text-motion consistency ("TM. Con."). The results are presented in Fig.~\ref{fig:userstudy}.
The findings reveal that GPT-4, without specialized prompt engineering, struggles to accurately capture joint interactions, global movement trajectories, and overall scene planning. In contrast, our method employs a more advanced tree-of-thought approach, which breaks down the planning process into group division, scene-guided, and event-driven stages. It allows us to effectively address the challenges of semantic and spatial integration, which GPT-4 struggles with, leading to improved performance in both crowd scene planning and motion generation.

\begin{table}[t]
\centering
\vspace{-1em}
\caption{\textbf{User studies for motion generation.}}
\vspace{-1em}
\label{tab:user-motion}
\resizebox{0.47\textwidth}{!}{
\begin{tabular}{l||ccc}

\toprule[1pt]
{} & $\textbf{B1}$ & $\textbf{B2}$ & Ours \\

{Alignment with spatial control signal} & $20.96\%$ & $13.87\%$ & $\textbf{65.17\%}$ \\
{Motion quality} & $9.05\%$ & $13.04\%$ & $\textbf{77.91\%}$ \\
\midrule[1pt]
\end{tabular}
}
\vspace{-2em}

\end{table}
\noindent\textbf{User studies for motion generation.}
Since existing techniques are unable to handle the task of crowd motion generation, we further compare our approach with two baseline methods: $\textbf{B1}$ - ``w/o ControlAttentio'' and $\textbf{B2}$ - ``w/o InputMixing'', both of which apply our proposed crowd scene planner. To evaluate this and differentiate from the previous user studies and volunteers, we invited another 25 volunteers, including investors, game designers, and gaming enthusiasts aged 25 to 40, to select the best results based on two criteria: (i) alignment with the spatial control signal, and (ii) motion quality. In practical, each volunteer was presented with 12 pairs of comparisons. The results, as shown in Tab.~\ref{tab:user-motion}, indicate that our design (1) enhances the controllability of the generated motion, leading to better alignment with the textual descriptions, and (2) produces motions of higher quality.

\subsection{Qualitative evaluations}

As illustrated in Fig.~\ref{fig:qualitative} (a)-(f), we randomly generated three distinct types of crowd motions using our method (\OM): 1) multi-person close interactions (a)-(b), 2) large-scale crowd dynamics under specific perturbations (c)-(d), and 3) complex crowd scene arrangements (e)-(f).
The generated motions exhibit natural and coherent motion semantics, realistic dynamics, and precise joint interactions. These results highlight that our proposed method excels not only in scene planning but also in generating realistic and accurate collective motions.

\subsection{Ablation studies}
We present the quantitative results of our ablation studies in Tab.~\ref{tab:ablation}. The following key observations can be made:
\textbf{1)} Comparing experiments (1) and (2) with our method, we find that both the ControlAttention and InputMixing components are effective in incorporating the spatial control signal into the motion transformer. This integration enables the model to generate high-quality motion sequences consistent with the provided textual description and spatial control signals.
\textbf{2)} Our inclusion of controlled keypoint loss ($\mathcal{L}_{con}$) and foot skating loss ($\mathcal{L}_{foot}$) significantly improves keypoint accuracy and reduces foot skating. This is demonstrated in experiments (4) and (5).
\textbf{3)} In experiment (7), we use only the ControlNet structure and inverse kinematics (IK) guidance to generate human motions, which simulates the method proposed by InterControl~\cite{wang2024intercontrol}. The results show a noticeable drop in motion quality and spatial consistency compared to our approach.
\textbf{4)} Finally, experiment (6) demonstrates that applying IK guidance during inference further enhances the overall quality of the generated motion.

\section{Conclusion}

In this work, we introduce the novel task of collective motion generation, which focuses on realistically generating crowd human motions tailored to specific scene contexts and events. To address this challenge, we propose \OM, a zero-shot, text-driven framework. Our approach begins by leveraging pre-trained crowd knowledge from large language models to categorize individuals into distinct groups and assign context-appropriate activities to each person. We then introduced a collective motion generator that integrates 3D spatial control signals within a transformer network, enhancing both the controllability and quality of the generated motion. Extensive evaluations demonstrate that our method produces promising results across various scenarios, marking a strong foundation for future advancements in this area.

\noindent\textbf{Limitations.}
\yk{While demonstrating the potential for collective motion generation, we recognize that \OMO has certain limitations:
1) Our crowd scene planner heavily relies on the large language model, which may struggle to accurately interpret complex or rare crowd scenarios. This could result in undesirable outcomes for human motion generation;
2) Currently, integrating scene information into the large language model remains challenging, which can hinder the creation of more realistic collective motion. As a result, the generated motions may show interactions with the physical scene when directly applied to real-world situations.}

\noindent\textbf{Potential Social Impacts.}
\yk{The proposed \OMO may offer significant benefits by enhancing realism and interactivity in virtual environments for entertainment and urban planning. However, it can also be misused to fabricate deceptive crowd scenes for simulations or entertainment, potentially misrepresenting public events or influencing opinions. 
}

\section*{Data availability statement}
\yk{The HumanML3D~\cite{guo2022generating}, KIT Motion Language (KIT-ML)~\cite{plappert2016kit} datasets are publicly available for research purpose - \href{https://github.com/EricGuo5513/HumanML3D}{HumanML3D}, \href{https://motion-database.humanoids.kit.edu/list/motions/}{KIT-ML}}



{\small
\bibliographystyle{plainnat}
\bibliography{reference}
}

\end{sloppypar}
\end{document}


\maketitle

\appendix

\appendix


\section{Details of CrowdMoGen}

\subsection{Diffusion Model}
\label{sec:diffusion}

Diffusion models can be defined using a Markov chain $p_{\theta}(\mathbf{x}_0)\,:=\,\int{p_{\theta}(\mathbf{x}_{0:T})\,d{\mathbf{x}_{1:T}}}$, where the intermediate variables $\mathbf{x}_{1},\cdots,\mathbf{x}_{T}$ represent noised versions of the original data $\mathbf{x}_0 \sim q(\mathbf{x}_0)$, and maintains the same dimensionality as the original data. In motion generation, every $\mathbf{x}_t$ corresponds to a motion sequence $\theta_i \in \mathbb{R}^D, i=1,2,\dots,F$, where $D$ denotes the dimensionality of each pose, and $F$ represents the total number of frames. In the diffusion process of diffusion models, ~\cite{ho2020denoising} efficiently derive $\mathbf{x}_t$ from $x_0$ by approximating $q(\mathbf{x}_t)$ as $\mathbf{x}:=\sqrt{\bar{\alpha}_t} \mathbf{x}_0 + \sqrt{1 - \bar{\alpha}_t}\epsilon$, where $\alpha_t := 1 - \beta_t$, $\bar{\alpha}_t:=\prod_{s=1}^t \alpha_s$, and $\beta_t$ is a predefined variance schedule at the $t^{\text{th}}$ noising step. 
Then, the diffusion process is reversed for clean motion recovery. Specifically, a motion generation model $S$ is configured to refine noisy data $\mathbf{x}_t$ back to its original, clean state $\mathbf{x}_0$. Following approaches like MDM~\cite{tevet2022human} and ReMoDiffuse~\cite{zhang2023remodiffuse}, we estimate $\mathbf{x}_0$ as $S(\mathbf{x}_t,t,c)$, where $c$ represents the conditioning signal containing the spatial control signal $c^s$ and the textual control signal $c^t$, and $t \in \mathcal{U}(0, T)$ marks the timestep in the diffusion process. Finally in the sampling phase, we derive $\mathbf{x}_{t-1}$ by sampling it from the Gaussian distribution $\mathcal{N}(\mu_{\theta}(\mathbf{x}_t,t,c), \beta_t)$, where the mean is specified as follows:
\begin{equation}
    \begin{aligned}
    & \mu_{\theta}(\mathbf{x}_t,t,c) = \sqrt{\bar{\alpha}_t} S(\mathbf{x}_t,t,c) + \sqrt{1 - \bar{\alpha}_t}\epsilon_{\theta}(\mathbf{x}_t,t,c), \\
    & \epsilon_{\theta}(\mathbf{x}_t,t,c)=(\frac{\mathbf{x}_t}{\sqrt{\bar{\alpha}_t}} - S(\mathbf{x}_t,t,c)) \sqrt{\frac{1}{\bar{\alpha}_t}-1}.
    \end{aligned}
\end{equation}

\subsection{Human Motion Representation. } 

Human motion can be parameterized using two primary methods: relative or global joint representation. The relative representation, proposed in ~\cite{guo2022generating}, is a redundant format that includes pelvis velocity, local joint positions, velocities, and rotations relative to the pelvis, complemented by foot contact binary labels. This representation method simplifies the model learning process and supports the generation of natural human motions aligning with inherent human body priors. However, it faces challenges in precise spatial control, especially in sparse time frames and when controlling joints beyond the pelvis, due to its lack of a stable global reference. Consequently, controlling methods~\cite{tevet2022human, karunratanakul2023guided} using this format struggle with consistency and precision in joint-wise spatial control across timesteps.
In contrast, global joint representation provides each joint’s absolute coordinates within a global framework, offering clear and direct reference for fine-grained control. Therefore, inspired by~\cite{xie2023omnicontrol}, we employ a hybrid approach to represent motions in our framework: utilizing relative representation for training and routine generation, and switching to global positioning for enhanced spatial control.

\subsection{IK Guidance}

Noise optimization during the sampling stage enhances the spatial control capabilities of our model. In diffusion steps, gradient descent optimization is performed on the output mean in response to the input spatial control signals. Specifically, $\mathcal{D}(\mu_{\theta}(\mathbf{x}_t,t,c), c^s)$ quantifies the discrepancy between the currently generated mean $\mu_{\theta}(\mathbf{x}_t,t,c)$ and the provided spatial signal $c^s$. To ensure precise control, we can selectively supervise specific time frames and joints, while masking unsupervised components. 
\begin{equation}
    \begin{aligned}
    \mathcal{D}(\mu_{\theta}(\mathbf{x}_t,t,c), c_s) = \frac{\sum_{i} \sum_{j} O_{ij} \mathcal{K}(c_{ij}^{s} - \mu^{g}_{ij})}{\sum_{i} \sum_{j} O_{ij}}
    \end{aligned}
\end{equation}
Here, $O_{ij}$ is a boolean value representing if at frame $i$, joint $j$ is controlled. The predicted mean is transformed into a global joint representation $\mu^g$ and assessed with the control signal $c^s$ in the same format under a pre-defined loss calculation function $\mathcal{K}$, suitable for different circumstances, ensuring that joint positions align with the spatial control requirements from a global perspective in both time and layout. Additionally, since gradient updates target the mean $\mathbf{x}_t$ in a relative motion representation, these updates are naturally propagated across all time frames and joints throughout the motion sequence. This coherent propagation facilitates a natural adjustment of the entire body’s joints, even when only a few joints are being actively controlled and edited.

\section{Details of Evaluation Metrics}
Detailed explanations of the metrics and the methods used are provided here due to article length constraints.

\paragraph{Semantic-Level Evaluation Metrics}
The following metrics are usually used to evaluate traditional text-to-motion task. To quantitatively assess, motion and text feature extractors are developed and trained under contrastive loss to produce feature vectors that are close geometrically for matching text-motion pairs and distant for non-corresponding pairs. \textbf{(1) R-Precision:} A pool is formed from the accurate description of a generated motion and 31 unrelated descriptions from the test dataset. Euclidean distances between the motion's features and those of each description in the pool are calculated and ranked. Accuracy is then averaged at the top-1, top-2, and top-3 levels, with successful retrievals marked by the ground truth description appearing within the top-k selections. \textbf{(2) MultiModality:} A distance is calculated by averaging the Euclidean distances between each generated motion's features and the text features of its respective description in the test set. \textbf{(3) Frechet Inception Distance (FID):} FID reflects the generation quality by calculating the distance between features extracted from real and generated motion sequences. \textbf{(4) Diversity:} Diversity measures the variability and richness of the generated motion sequences. \textbf{(5) Multi-Modality:} Multi-modality measures the average variance of different generated motion sequences given a single text description.
We use R-Precision, FID, and Diversity for evaluating our method on the text-to-motion task, since these metrics can best showcase the generation quality and control ability of a model.

\paragraph{Spatial-Level Evaluation Metrics}: Metrics such as Foot skating ratio, Trajectory error, Location error, and Average error of keyframe locations are employed to evaluate our proposed method's spatial control ability.  \textbf{(1) Trajectory error} refers to the proportion of trajectories that are unsuccessful, which occurs when any keyframe's location error exceeds a set threshold. \textbf{(2) Average error} is computed as the mean distance between the positions of generated motions and the actual keyframe positions observed during the key motion steps. \textbf{(3) Location error} represents the ratio of keyframe locations that are not fall within a threshold distance.  \textbf{(3) Foot skating ratio} is a metric used to quantify the amount of sliding or unrealistic movements of feet in animations.

\section{Limitations.}
While demonstrating the potential for collective motion generation, we recognize that \OMO has certain limitations:
1) Our crowd scene planner heavily relies on the large language model, which may struggle to accurately interpret complex or rare crowd scenarios. This could result in undesirable outcomes for human motion generation;
2) Currently, integrating scene information into the large language model remains challenging, which can hinder the creation of more realistic collective motion. As a result, the generated motions may show interactions with the physical scene when directly applied to real-world situations.

\section{Potential Social Impacts.}
The proposed \OMO may offer significant benefits by enhancing realism and interactivity in virtual environments for entertainment and urban planning. However, it can also be misused to fabricate deceptive crowd scenes for simulations or entertainment, potentially misrepresenting public events or influencing opinions. 

{\small
\bibliographystyle{plain}
\bibliography{reference}
}